\documentclass[journal]{IEEEtran} % use the `journal` option for ITherm conference style
\IEEEoverridecommandlockouts
\usepackage{cite}
\usepackage{amsmath,amssymb,amsfonts}
\usepackage{algorithm}
\usepackage{algpseudocode}
\usepackage{graphicx}
\usepackage{textcomp}
\usepackage{xcolor}
\usepackage{booktabs}
\usepackage{url}
\usepackage{multirow}
\def\BibTeX{{\rm B\kern-.05em{\sc i\kern-.025em b}\kern-.08em
    T\kern-.1667em\lower.7ex\hbox{E}\kern-.125emX}}
\begin{document}

\title{Two-Step Reinforcement Learning for Multistage Strategy Card Game}

\author{%%%% author names
    \IEEEauthorblockN{Konrad Godlewski},
    \IEEEauthorblockN{Bartosz Sawicki},
    % duplicate the line above as many times as needed to list all authors
    \\%%%% author affiliations
    \IEEEauthorblockA{\textit{Warsaw University of Technology}},\\% first affiliation
    % duplicate the line above as many times as needed to list all affiliations
    %%%% corresponding author contact details
    \IEEEauthorblockA{jowisz11@gmail.com, bartosz.sawicki@pw.edu.pl}
}

\maketitle

\begin{abstract}
In the realm of artificial intelligence and card games, this study introduces a  two-step reinforcement learning (RL) strategy tailored for "The Lord of the Rings: The Card Game (LOTRCG)," a complex multistage strategy card game. This research diverges from conventional RL methods by adopting a phased learning approach, beginning with a foundational learning stage in a simplified version of the game and subsequently progressing to the complete, intricate game environment. This methodology notably enhances the AI agent's adaptability and performance in the face of LOTRCG's unpredictable and challenging nature. The paper also explores a multi-agent system, where distinct RL agents are employed for various decision-making aspects of the game. This approach has demonstrated a remarkable improvement in game outcomes, with the RL agents achieving a winrate of 78.5\% across a set of 10,000 random games. 
\end{abstract}

\begin{IEEEkeywords}
    reinforcement learning, incremental learning, card game.
\end{IEEEkeywords}

\section{Introduction}

%%%%%%%%%%%%%%%%%%%%%%%%%%%%%

Card games have experienced rapid growth in recent years, as proved by the increasing number of titles on platforms such as Google Play, App Store or Steam. This popularity is largely based on features such as deck building, short gaming sessions and replayability, which results from random events during gameplay. All this means that card games can provide hours of entertainment for their lovers. 

Modern artificial intelligence methods have proven to outperform humans in many tasks, however card games containing a high degree of randomness are still a challenge. For example, the first AI agent capable of beating professional players of no-limit Texas Hold'em was not created until 2019 \cite{brown2019superhuman}. Collectible Card Game (CCG) is class of the card games which allows the decks of cards to be redefined, what requires significant adaptation from the player. 
The "Lord of the Rings: Card Game" is a representative of this type of card games, what makes it interesting research object. 

The AI algorithms detailed in this work fall into the category of Reinforcement Learning (RL), a branch of machine learning that relies on a trial-and-error approach. The learning process involves the interaction of an agent with its environment, where the agent observes the game state and decides on appropriate actions. These actions are implemented in the environment, which provides a reward in return. There are numerous examples of the implementation of RL techniques in card games. 

Three RL algorithms: Deep Q-Learning, A2C, and PPO were compared in the game \textit{Chef's Hat} (Barros et al. \cite{barros2021learning}). \textit{Chef's Hat} is a competitive four-player card game in which players try to become a chef. The game is played in turns, in which each player decides whether to play cards or fold. Due to the four-player nature of the game, it was possible to experiment with different configurations of agents. RL algorithms were challenged with random agents in direct skirmishes and against a human player.

Proximal Policy Optimization (PPO) algorithm was applied to the \textit{DouDizhu} game (Guan et al. \cite{yang2022perfectdou}). \textit{DouDizhu} is a three-player game mixing collaboration with competition. Two players have to cooperate in order to defeat the third player. The authors developed a framework called Perfect-Training-Imperfect-Execution (PTIE) based on centralized training and decentralized execution of RL agents. PTIE allows agents to train their policies on a game that is treated as perfect information. The policies are distilled in order to play the actual game with imperfect information.

Usage of PPO was investigated for a drafting phase in collectible card games (Vieira et al. \cite{vieira2020drafting}). The methods built a deck in three variants that differ in the representation of the game state. The first variant based on the MLP network includes all previously selected cards into the state vector. In the second, the leading role is played by the LSTM network, which accumulates information about previous card choices only based on the vector of cards currently available to the player. Finally, the third option uses only the MLP network with the representation the same as in the second option.

Zha et al. \cite{zha2019rlcard} developed an open-source platform to learn and test reinforcement learning agents on card games. The platform supports standard 52-card games like Blackjack, Texas Hold'em, and also Chinese-originated games such as Mahjong and \textit{DouDizhu}. The authors test three algorithms on their platform, such as Deep Q-Network, Neural Fictitious Self-Play, and one outside RL such as Counterfactual Regret Minimization.

In order to accelerate learning process in Mahjong a mirror loss function was proposed \cite{zhao2022improving}. It allows the RL agent to take mirrored actions in the mirrored environment. This method limits the policy space during the optimization process.

Yao et al. \cite{yao2022towards} propose a method of handling large action spaces of the \textit{Axie Infinity} card game. \textit{Axie Infinity} is an online competitive 2-player game in which the player forms a few subsets of cards in order to defeat the opponent. Cards to a subset are chosen based on Q-function approximation. The function indicates the optimal decision from a restricted subset of actions. The method can be applied to other card games with large action spaces like \textit{DouDizhu} or \textit{Hearthstone}.

State representation in the 52-card game of Hearts was Sturtevant and White \cite{sturtevant2007feature} analyzes the performance of the RL agent for different state representations in the game of Hearts.

The Hanabi card game has been seen as a challenge to AI in recent years \cite{bard2020hanabi}. Hanabi is a cooperative card game for 2-5 players. The game stands out for the way it handles imperfect information. The player does not see his own cards; he can only observe other players' cards. In order to play the right card, the player must get hints from other participants.

Recently Hanabi has drawn the attention of RL researchers. Grooten \cite{grooten2022vanilla} et al. compare different RL algorithms for Hanabi. The algorithms include Proximal Policy Optimization (PPO), Vanilla Policy Gradient (VPG), and Simple Policy Gradient (SPG). VPG is an actor-critic method. It maintains separate neural networks for both policy and value function approximation. SPG uses only a policy network. The authors analyze various aspects of the algorithm's performance within the game, such as learning curves and policy refinement over episodes.

Other noteworthy applications of artificial intelligence algorithms in card games can be found in the following papers: Hanabi \cite{canaan2022generating}, Splendor \cite{bravi2022rinascimento}, and Leduc Hold'em \cite{guo2023suspicion}. Kowalski and Miernik \cite{kowalski2020evolutionary} put forward the use of an evolutionary algorithm for card drafting in the game \textit{Legends of Code and Magic}.

The primary innovation of this paper is its strategy for implementing a two-step reinforcement learning (RL) technique in the intricate and random  world of LOTRCG. The methodology tackles the game's inherent complexity by first learning the RL agent in a straightforward environment and then advancing to the complete level of difficulty. Exploring a multi-agent setup where various RL agents specialize in distinct decision phases of the game represents a significant advancement in AI-driven strategy gaming.

The paper comprises four main sections.  Section 2. Reinforcement Learning Agent explores how the actor-critic model, state encoders, and action decoders are applied in LOTRCG. The study progresses to 3. Learning Strategies, which compares various approaches, such as one-step learning, two-step learning, and interrupted learning, and highlights their effectiveness in the game's challenging environment. A distinctive feature of this paper is the 4. Multi-Agent Setup section, in which the efficacy of single versus multiple RL agents during decision-making phases is examined. The paper concludes with 5. Conclusions, which recap the findings and highlight the effectiveness of the two-step learning approach.

\section{Multistage game}

The Lord of the Rings: The Card Game (LOTRCG) is a fantasy-themed, cooperative, collectible card game published in 2011 by Fantasy Flight Games. The game is based on challenging adventures in order to complete a scenario. During the scenario inspired by famous J.R.R.~Tolkien universe, a fellowship led by the players encounters many adversities, and if they fail, the scenario is lost. The game can be played in two variants: solo or two-player cooperative. In LOTRCG, the players can form their fellowship using a variety of decks. By default, the core set features four decks, but a more appealing option is building its own.  
The game is receiving positive reviews in the card game community, however they recognize its steep learning curve.

\begin{figure}
\centering
\includegraphics[width=0.5\textwidth]{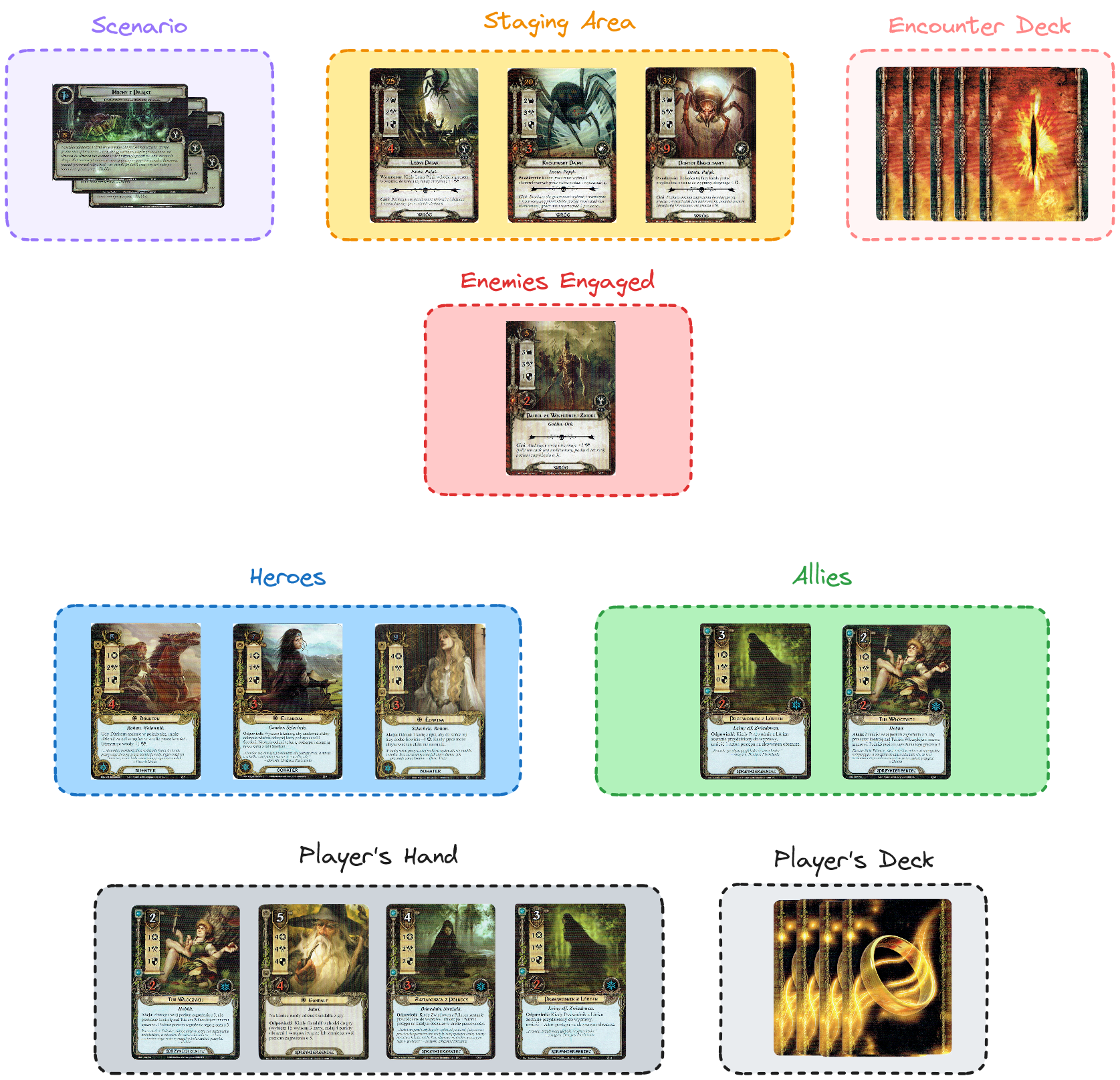}
\caption{An example game state at the planning phase.}
\label{fig:game_state}
\end{figure}

The game's goal is to complete a scenario, which means reaching a specified number of progress points throughout the game. During the scenario, the player encounters objects such as enemies or lands, which hinder gaining progress points. These objects require a player's reaction. If an enemy appears, the player has to defend himself; otherwise, he can lose his heroes or allies. If a land card appears, the player can decide whether explore it. Leaving lands unexplored makes the scenario progress difficult.

\begin{figure}
\centering
\includegraphics[width=0.5\textwidth]{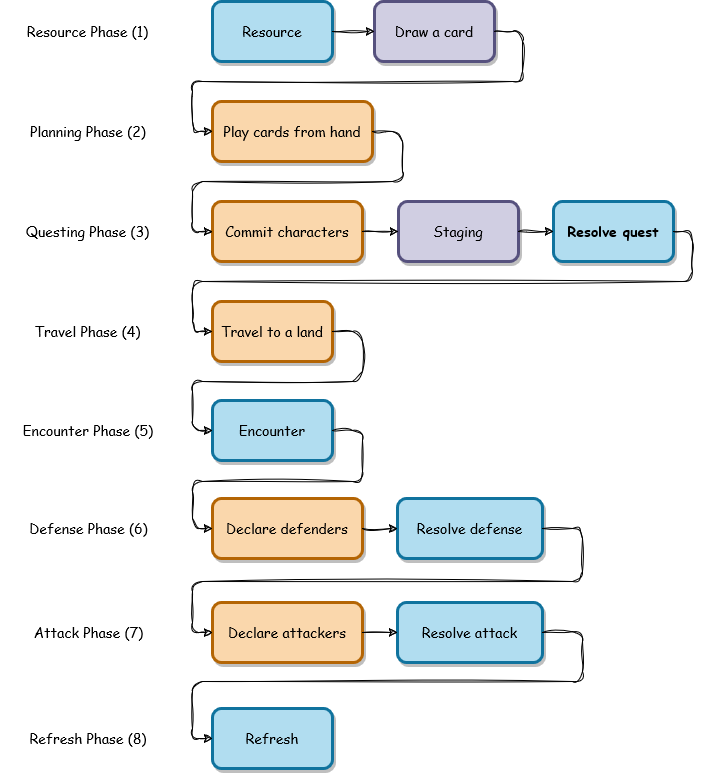}
\caption{The sequence of activities that constitute each phase. Activities are ordered in rule-based (blue), random events (violet) and player decisions (orange). The bolded rect marks the moment of determining game-end conditions.}
\label{fig:phase_seq}
\end{figure}

A game round consists of eight phases (Fig.~\ref{fig:phase_seq}), starting with the resource phase where players draw cards and gain resources, and followed by the planning phase for card purchases. During the questing phase, players commit characters to quests and face threats, and in the travel phase, they can explore lands. The encounter and defense phases involve engaging in fights with enemies, and players counterattack in the attack phase. The refresh phase readies all characters and increases the player's threat level. The game's outcome is determined during the questing and defence phases, where players win by achieving quest points or lose if their threat level exceeds 50 or all heroes die.

The game difficulty could be controlled by the number of points required for game success. The number 20 is the default threshold specified by the game rule book. However, for the purposes of this article, the difficulty was varied from 8 to 20 points.

\section{Reinforcement Learning Agent}

Reinforcement Learning (RL) has been a new trend in the development of artificial intelligence in recent years. The concept is based on the trial-and-error method~\cite{sutton2018reinforcement}, in which an AI agent interacts with an environment. The AI agent is a decision-making algorithm that takes specific actions based on observations. The environment is the entity where this action will be executed, and a feedback signal (positive, negative, or zero) is sent to the agent. RL differs significantly from other machine learning techniques. RL agent does not operate on a static set of learning data but receives a feedback signal based on which it performs the learning process. The feedback signal itself is irregular, which means positive or negative information may appear at different time intervals. These features are ideally suited to the field of games, where it may not be possible to generate a sufficiently large set of learning data. The win/loss feedback signal also occurs at different times since one game round ends after ten turns and another after four.

The basic concepts used in the context of reinforcement learning are as follows:
\begin{itemize}
\item State $(s)$- current game situation including all information coming from the table and hand of the player (see~Fig.~\ref{fig:game_state}),
\item Action $(a)$ - game action resulting from a decision-making process,
\item Policy $\pi(s)$ - a function that maps state probability distribution over actions,
\item Reward $(r)$- environment's reaction to action,
\item Value Function - represents a measure of how beneficial it is for a player to be in a given state or state-action pair. Value function for policy $\pi$ can be described with the following equations \cite{sutton2018reinforcement}:
\begin{equation}
\label{eq:valueV}
v_{\pi}(s) = \displaystyle\sum_a \pi(a|s) \displaystyle\sum_{s',r}p(s',r|s,a) [r + \gamma v_{\pi}(s')]
\end{equation}
\begin{equation}
\label{eq:valueQ}
q_{\pi}(s,a) = \displaystyle\sum_{s',r}p(s',r|s,a) [r + \gamma q_{\pi}(s',a')]
\end{equation}
where: $v_{\pi}(s)$ - value function for state $s$, $\pi(a|s)$ - probability of action $a$ in state $s$, $p(s',r|s,a)$ - probability of reaching next state $s'$ and reward $r$ by performing action $a$ from state $s$, $\gamma$ - discount factor, $v_{\pi}(s')$ - value function for the next state $s'$, $q_{\pi}(s,a)$ - value function for the pair state $s$ and action $a$, $q_{\pi}(s',a')$ - value function for the next state $s'$ and the next action $a'$.
\end{itemize}

The goal of reinforcement learning is the maximization of discounted reward in the long term \cite{sutton2018reinforcement}:
\begin{equation}
\label{eq:RLsum}
Q_{n+1} = \frac{1}{n} \displaystyle\sum_{i=1}^{n} R_{i} = Q_{n} + \frac{1}{n} [R_{n} - Q{n}]
\end{equation}
This formula allows us to iteratively calculate the value function $Q_{n+1}$ given its current value $Q_n$ and reward $R_n$. This goal is achieved through Generalised Policy Iteration (GPI). GPI consists of two steps: calculating the value function and modifying the strategy (policy improvement). The value function can be calculated from the formulas (\ref{eq:valueV}) or (\ref{eq:valueQ}) for each state and action for low complexity problems. However, approximation methods such as linear or neural networks are employed for large state spaces. Once the value function is obtained, the strategy is updated according to the formula:
\begin{equation}
\label{eq:policyImprovement}
\pi(s) \leftarrow \operatorname*{argmax}_a \displaystyle\sum_{s',r}p(s',r|s,a) [r + \gamma v_{\pi}(s')]
\end{equation}

The type of RL algorithm used in this paper is Actor-Critic (AC). It approximates both the value function and the strategy. The actor is responsible for estimating the probability distributions of actions for a given strategy according to the equation \cite{sutton2018reinforcement}:

\begin{equation}
\label{eq:ACpolicy}
\pmb{\theta}_{t+1} \leftarrow \pmb{\theta}_{t} + \alpha [R_{t+1} + \gamma v(S_{t+1}, \pmb{w}_{t}) - v(S_{t}, \pmb{w}_{t})] \frac{\nabla \pi(A_{t}|S_{t}, \pmb{\theta}_{t})}{\pi(A_{t}|S_{t}, \pmb{\theta}_{t})},
\end{equation}
where $\pmb{\theta}$ and $\pmb{w}$ are vector parameters of policy and value function respectively. The critic is responsible for the value function as follows:

\begin{equation}
\label{eq:ACstatevalue}
\pmb{w}_{t+1} \leftarrow \pmb{w}_{t} + \alpha [R_{t+1} + \gamma v(S_{t+1}, \pmb{w}_{t}) - v(S_{t}, \pmb{w}_{t})] \nabla v(S_{t}, \pmb{w}_{t}).
\end{equation}

The adaptation of the RL learning model to LOTRCG required implementation of several modules such as the underlying agent and the environment along with the auxiliary classes like the simulator and encoders. The simulator class handles the exchange of information between the agent and the environment. The observation from the environment is passed to the agent, which chooses the action. The simulator applies this action to the environment, which returns reward and next observation, which along with current observation and action form an input vector to the agent. The agent performs a learning process upon that input vector.

The above description is common to all implementations of the RL algorithm. What distinguishes the different problems is the communication scheme between the environment and the agent. These are necessary to handle the input and output of the neural network. This issue is described in the subsections on state encoding and action decoding.

\subsection{State Encoders}

Encoders handle the data flow from the environment to the agent. They fetch data from the game model and embed it in a feature vector. This vector then feeds the agent as a neural network input. The process is performed for planning and questing separately.

\begin{figure}
\centering
\includegraphics[width=0.5\textwidth]{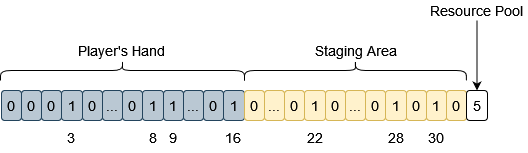}
\caption{Example of the encoding scheme for the planning phase presented in Fig.~\ref{fig:game_state}. Cards in the player's hand are: Lorien Guide (id:3), Northern Tracker (8), Wandering Took (9) and Gandalf (16). There are three enemies in the staging area: Forest Spider (22), King Spider (28) and Ungoliants Spawn (30). The last element of the vector is set to the current resource pool of 5 points. }
\label{fig:planning_encoding}
\end{figure}

The feature vector for planning phase consists of 29 binary values (Fig.~\ref{fig:planning_encoding}):
\begin{itemize}
    \item 17 binary vector defining ally cards in the player's hand,
    \item 15 binary vector defining enemy cards in the staging area,
    \item an integer specifying total resource pool available to the player.
\end{itemize}

The feature vector for the questing phase also relies on binary and integer variables, but it points to different cards and statistics regarding a situation on the board as follows:
\begin{itemize}
\item Encoding type 0 - enemies in the staging area and round number,
\item Encoding type 1 - lands, enemies in the staging area and round number,
\item Encoding type 2 - enemies in the staging area and combined threat,
\item Encoding type 3 - enemies in the engagement area and combined threat.
\end{itemize}
Every vector has 18 binary variables representing the player's cards: three for hero cards and 15 for allies. The remaining features depend on a particular encoding scheme. The encoding for the defense decision consists of a binary vector with ids of hero, ally, and enemy cards.

We investigated the effect of game state encoding type on efficiency in the questing phase. Within four encodings the best winrate achieved the type 2 (92.2\%) followed by the type 3 (80.7\%). Opposite to remaining encodings (type 0 and 1), these two observe the combined threat, which is a sum of threat of all cards in the staging area (yellow rectangle in Fig.~\ref{fig:game_state}). The importance of the combined threat indicates that the RL agent is learning the game based on a condensed observation since the combined threat is a composite value that depends on the cards in the staging area, either lands or enemies.

\subsection{Action Decoder: Macroactions}

Action decoders serve as middle-ware between the agent and the environment. They receive an action from the agent and translate it into an executable form suited for the environment. In presented experiments two forms of action are analysed: a macroaction (abstract) or a direct card choice.

Macroactions offer level of abstraction instead of picking exact cards. It allows the agent to operate on a fixed number of actions, making the size of decision space  constant. The drawback of this approach is losing a degree of freedom of choice during the decision process.

Proposed macroactions scheme is based on an idea to introduce coefficient $\beta$, which could be understood as a choice of whether the player's strategy is to be more offensive or defensive. Weighting coefficient $\beta$ takes values from 0 to 1 with 0.2 step. 

For given value of $\beta$ the cards $c$ are sorted in descending order according to value of $f$ function:

\begin{equation}
f(c) = \frac{\beta * c_w + (1 - \beta) * c_d}{c_c} 
\label{eq:planning_formula}
\end{equation}
where $n$ is a number of available cards, $\beta$ - a weighting coefficient, $c_w$ - card willpower, $c_d$ - card defense and $c_c$ is card cost. 

When the ordering process is done, for planning phase cards are acquired until the total resource pool is depleted. 

For the questing phase, a similar formula applies. Cards are committed to the quest up to the combined threat level of the staging area.

\subsection{Action Decoder: Direct Card Choice}

\begin{algorithm}
\caption{Direct Card Choice - Planning}
\label{alg:planningPhase}
\begin{algorithmic}
\Require $agent, env$
\State $substate \gets env.encodeStatePlanning()$
\While{$len(substate.cardsAvailable) != 0$}
    \State $cardId \gets agent.actPlanning(substate)$
    \State $env.applyPlanning(cardId)$ 
    \State $substate \gets env.encodeStatePlanning()$
\EndWhile
\end{algorithmic}
\end{algorithm}

\begin{figure}
\centering
\includegraphics[width=0.45\textwidth]{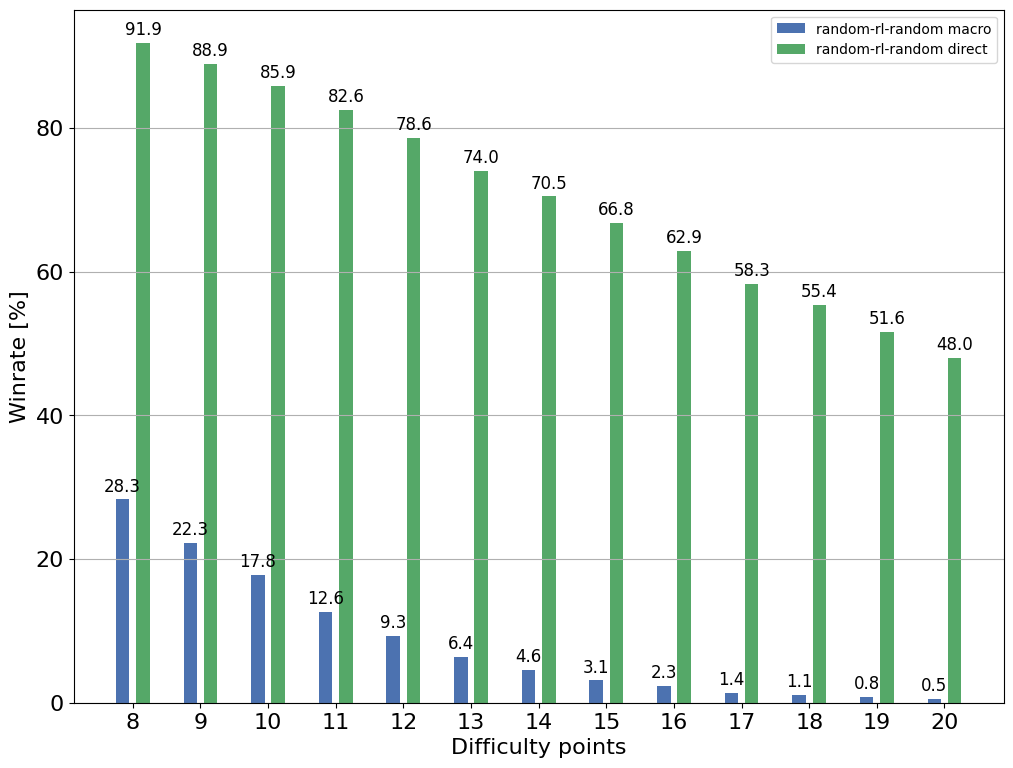}
\caption{Testing results of macro and direct agent setups learned at 8 difficulty points.}
\label{fig:difficulties}
\end{figure}

The direct method restricts the agent's response to a two-point distribution of whether a card should be played or not. These decisions are based on a planning substate. The decisions are executed in a loop until there are no cards affordable for the agent (Alg.~\ref{alg:planningPhase}). The loop passes the current substate to the agent, which returns an action. The action has a form of id of a card from the player's hand. The last step of the loop is applying to the action to the environment. 

\begin{table}
\centering
\caption{An example planning action loop.}
\begin{tabular}{ccccc}
\toprule
Hand & Staging Area & Res. Pool & Action & Table \\
\midrule
3, 8, 9, 16 & 22, 28, 30 & 5 & Play 9 & - \\
3, 8, 16 & 22, 28, 30 & 3 & Play 3 & 9 \\
8, 16 & 22, 28, 30 & 0 & - & 3, 9 \\
\bottomrule
\end{tabular}
\label{tab:planning_action}
\end{table}

Table ~\ref{tab:planning_action} presents an example execution of the loop. The agent has four cards in hand (ids: 3, 8, 9, 16) and five tokens in the resource pool. There are three enemies in the staging area (ids: 22, 28, 30). He decides to play a card (id:9) with a cost of 2 - it shows up on the table, and the resource pool gets updated. Then the agent purchases a card (id:3) with a cost of 3. The resource pool drops to zero, and the program breaks the loop. The agent's cumulative action consists of cards with id:9 and id:3 played in those two iterations.

\begin{algorithm}
\caption{Direct Card Choice - Questing }
\label{alg:questingPhase}
\begin{algorithmic}
\Require $agent, env$
\State $state \gets env.encodeStateQuesting()$
\State $cardIds \gets agent.actQuesting(state)$ 
\State $env.applyQuesting(cardIds)$
\end{algorithmic}
\end{algorithm}

Direct actions at the questing phase are processed in a no-looping workflow (Alg.~\ref{alg:questingPhase}). It begins with encoding a game state, taking action on that state. The action is a list of ids of available cards - hero and allies. Then the action is applied to the environment. Following the example, now the agent has three heroes (ids: 0, 1, 2) accompanied by four allies (ids: 3, 5, 9, 10). There are three enemies (ids: 22, 28, 30) in the staging area. The agent decides to commit (ids: 1, 3, 10) to the quest, leaving the rest (ids: 0, 2, 5, 9) for later phases such as the defense or attack phase (phase 6 and 7 in Fig~\ref{fig:phase_seq}).

\subsection{Hyper-parameter optimization}

The optimisation process of a neural network hyper-parameters in Actor-Critic model was performed.

\begin{table}[ht]
\centering
\caption{Winrate for setup: direct AC agent at planning, direct AC at questing and random agent at defense. Difficulty 8 points. }
\label{tab:rllow1}
\begin{tabular}{ccccc}
\toprule
number of neurons & learning rate & encoding & winrate \\
\midrule
70 & 6e-4 & 2 & $91.9 \pm 0.5$ \\
100 & 8e-4 & 2 & $88.7 \pm 0.6$ \\
70 & 7.5e-4 & 3 & $83.7 \pm 0.7$ \\
\bottomrule
\end{tabular}
\end{table}

Table \ref{tab:rllow1} presents three  optimized networks, which achieved the best results. Two of them used the questing state encoding type of 2, which allowed the winrate about 90\%. 

Networks with a minimum of 70 neurons in the hidden layer are effective. Increasing the number of neurons may cause overfitting, where the network only remembers specific actions for given observations instead of generalizing them. A high number of neurons can also lead to instability during the learning process.

Hyper-parameter optimization consisted of 100 trials. Each of them samples the search space, which means that the AI setup plays 10000 episodes in order to learn the RL agent. The evaluation function tracks the best average reward from recent 1000 episodes and records it as the trial score.

\section{Learning strategies}

Initial attempts to learn the model to play at the highest full difficulty level (20) were unsuccessful. Simply changing the values of the hyperparameters and increasing the computational budget had no effect. Hence, a step-by-step learning strategy was developed, inspired by the concepts of Curriculum Learning or Progressive Learning. 

Curriculum Learning (CL) in machine learning mimics the human educational approach of progressing from simple to complex concepts. Pioneered by Bengio et al. in 2009, CL structures the training of models by initially presenting easier tasks or examples, and gradually increasing complexity~ \cite{bengio2009curriculum}. This method has proven effective across various domains, including object localization, detection, and neural machine translation.
Progressive Reinforcement learning was applied to recognize actions in skeletal animation \cite{tang2018deep}. The method samples frames of the video and then progressively selects them based on their relevance in the animation sequence. 

Progressive Reinforcement Learning (PRL) has demonstrated significant potential in various domains, notably in recognizing actions in skeletal animation as illustrated by Tang et al. \cite{tang2018deep}. Their method effectively samples frames from videos, progressively selecting the most relevant ones in the animation sequence, showcasing the adaptability of PRL in handling sequential data.

An RL agent learned on a simple environment can be used for a new, more complex problem \cite{madden2004transfer}. This progressive technique consists of two sequential phases. In the first called experimentation the agent solves a simple problem using vanilla Q-Learning. Then introspection is performed, which generates a symbolic representation of the solved problem. This representation will then be used to gain knowledge of states unexplored by the agent in the next phase of experimentation already with an increased level of difficulty. The use of prior knowledge, gained from solving simple problems, was also presented for classification tasks such as image and audio recognition \cite{fayek2020progressive}.

The solution presented in this paper can be classified as a example of incremental learning. The research described in this section is based on a single RL agent setup, which will be extended in the next section. 

The learning process of a neural network is stochastic in nature. This is due to the random initialisation of the weights of the network but is mainly due to the strongly random nature of the LOTRCG game. There are two card draw events in each round of the game (violet rects in Fig.~\ref{fig:phase_seq}). This results in a learning process that is never repeatable.

\begin{figure*}
\centering
a) \includegraphics[width=0.3\textwidth]{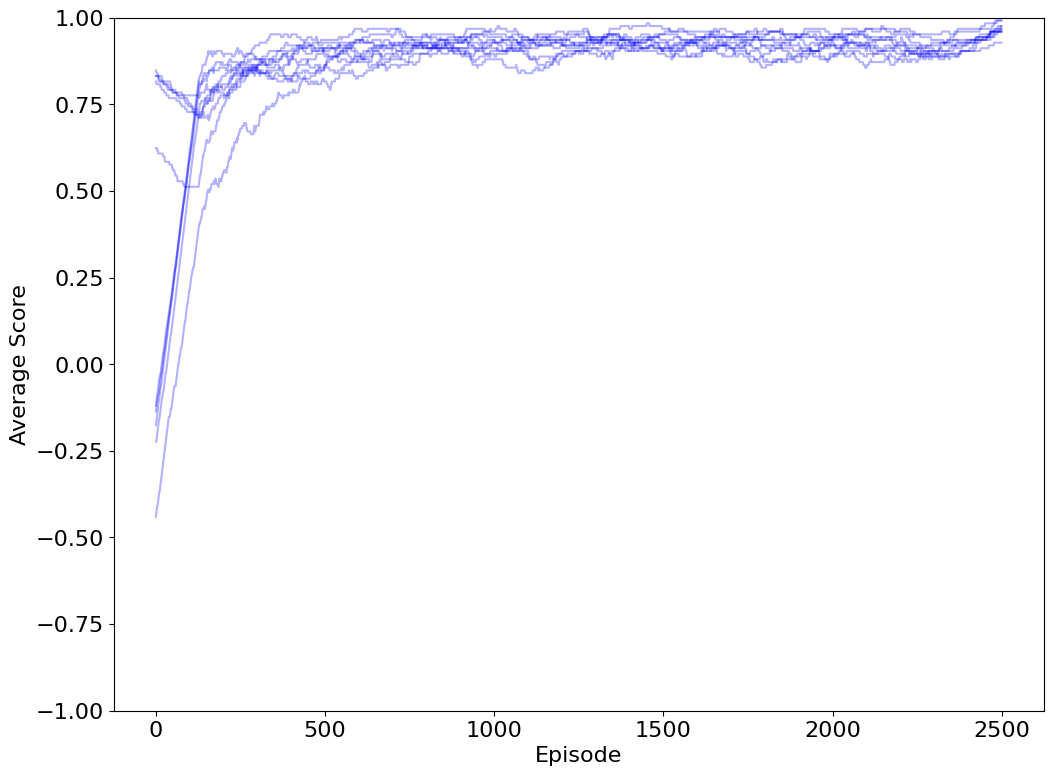}
b)\includegraphics[width=0.3\textwidth]{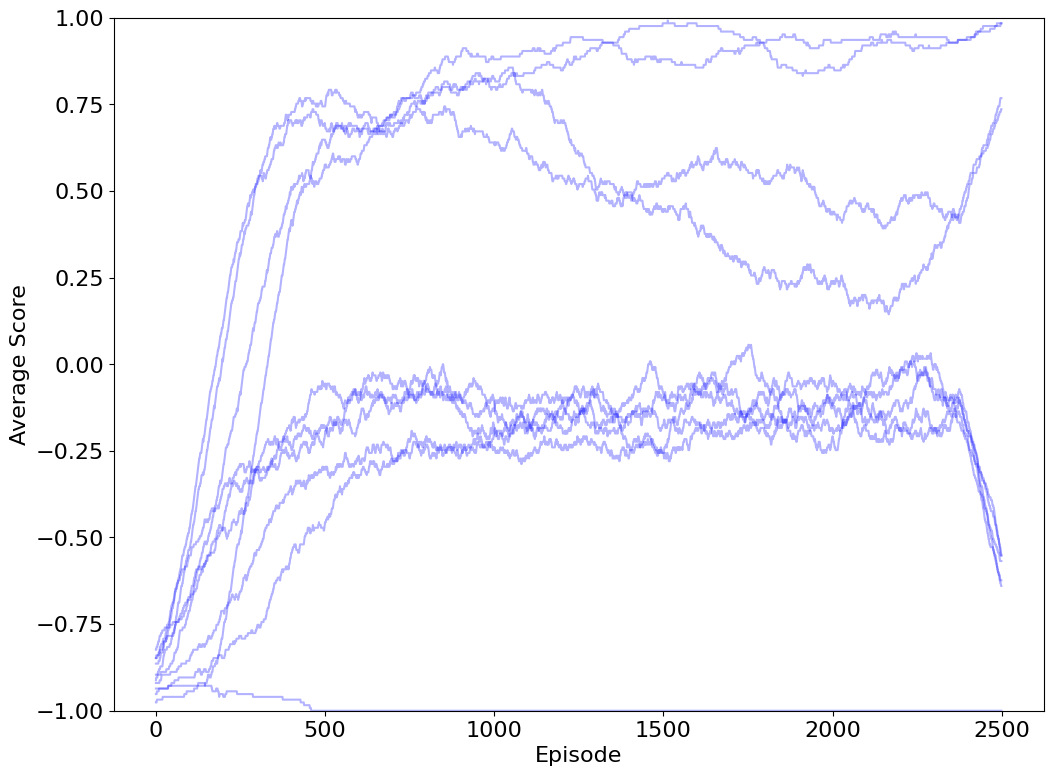}
c) \includegraphics[width=0.3\textwidth]{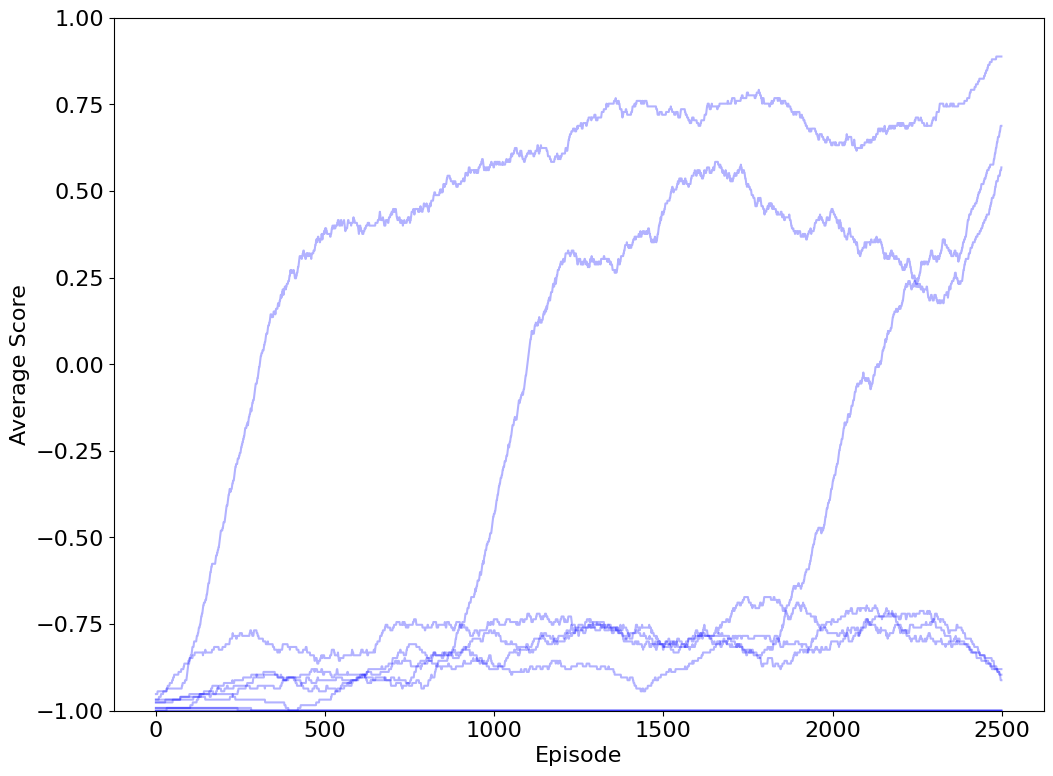}
\caption{Learning curves for three different game difficulty levels.  a) 1 point, b) 5 points and c) 9 points. The difficulty of the full game is 20 points.}
\label{fig:learning_curves}
\end{figure*}

The game's difficulty affects the model's ability to learn to win. Figure~\ref{fig:learning_curves} shows what 10 example learning curves look like for three selected difficulties of 1, 5, and 9 points. For the lowest difficulty (1 point), it only takes 500 episodes (full games) for the network to reach a very high win rate. For a game with a difficulty of 5 points (Fig.~\ref{fig:learning_curves}b), we observe large differences between the learning processes. In most cases, after 2500 episodes, the average reward is at an average level. However, it should also be noted that in a few cases, the learning process went practically perfectly. Playing at a difficulty level of 9 points is even more challenging. As shown in Fig.~\ref{fig:learning_curves}c, only a few learning processes can lead the network to positive results. Further raising the difficulty further strengthened this effect, and for a game with a difficulty of 20 points, not a single win was observed even after 20000 episodes.

%%%%%%%%%%%%%%%%%%%%%%%%%%%%%%%%%%%%%%%%%%
\subsection{One-step learning}

\begin{figure}[htbp]
\centering
\includegraphics[width=0.95\linewidth]{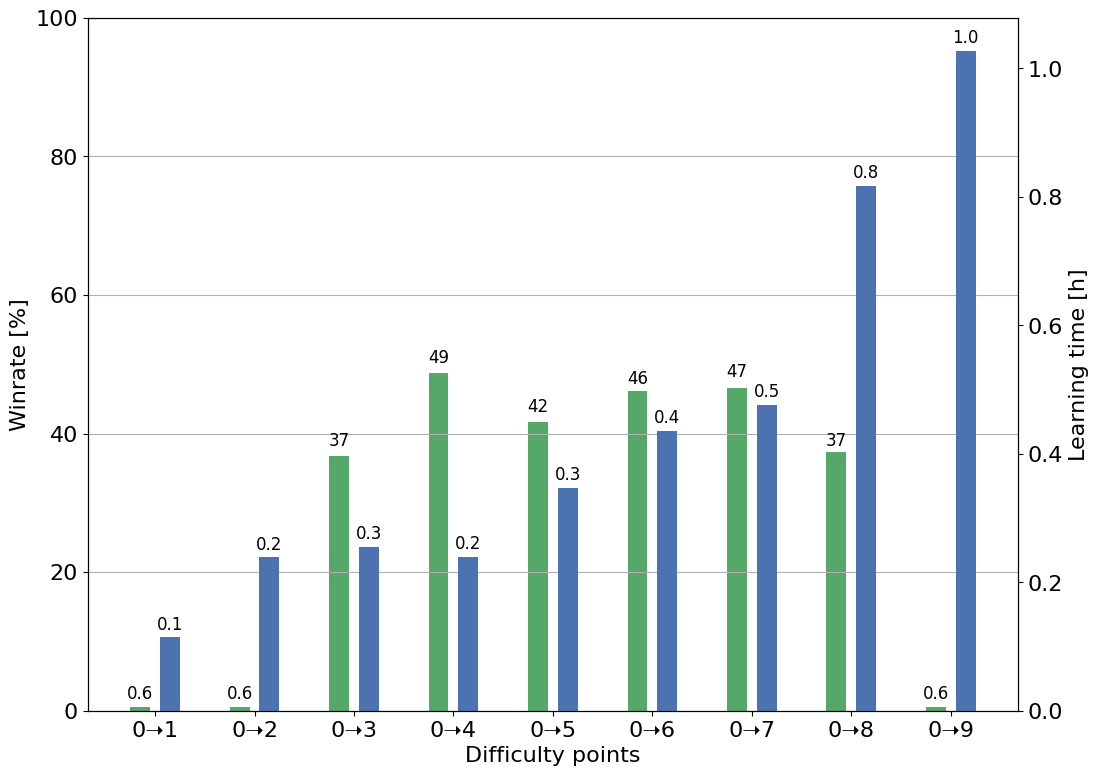}
\caption{Winrate of networks learned on the game with lower difficulties and tested on full difficulty (20 points).}
\label{fig:onestagetesting}
\end{figure}

If learning a network on a game of difficulty 20 has proved impossible, using networks learned on lower difficulties is the simplest solution. While these networks show some playability on a simplified game, they may also prove useful in a full-difficulty game. The results of this experiment are shown in Fig.~\ref{fig:onestagetesting}. Learning was carried out on nine variations of game difficulty. To improve reliability, ten iterations of the learning process of 1000 episodes were performed on each. Each network was then tested on a game with a difficulty of 20. It can be seen that only networks learned on difficulties 3 to 8 achieve a win rate above 30\%. The best result of 48\% was achieved by the network trained on difficulty 4.

%%%%%%%%%%%%%%%%%%%%%%%%%%%%%%%%%%%%%%%%%%
\subsection{Two-step learning}

If single learning has proven unsuccessful, one alternative is to split the learning process into two stages. During the first stage, the network would learn on a simplified difficulty, and in the second stage, it would learn on the full difficulty. However, the query remains as to the optimal threshold for this type of division. Efficient computation is crucial, considering that the network learning tasks in the RL model are highly time-consuming. One experiment may take several days to complete.

For a complete two-step learning, there are nine attempts at learning the network with reduced difficulty, followed by the learning process at full difficulty for each attempt. Due to the extensive computational time required, which would have taken many days, we abandoned this strategy and instead focused on devising more efficient solutions.

Therefore, the aim is to reduce the number of learning processes and shorten their length while keeping the winrate as high as possible. 

\begin{figure}[htbp]
\centering
a)\includegraphics[width=0.8\linewidth]{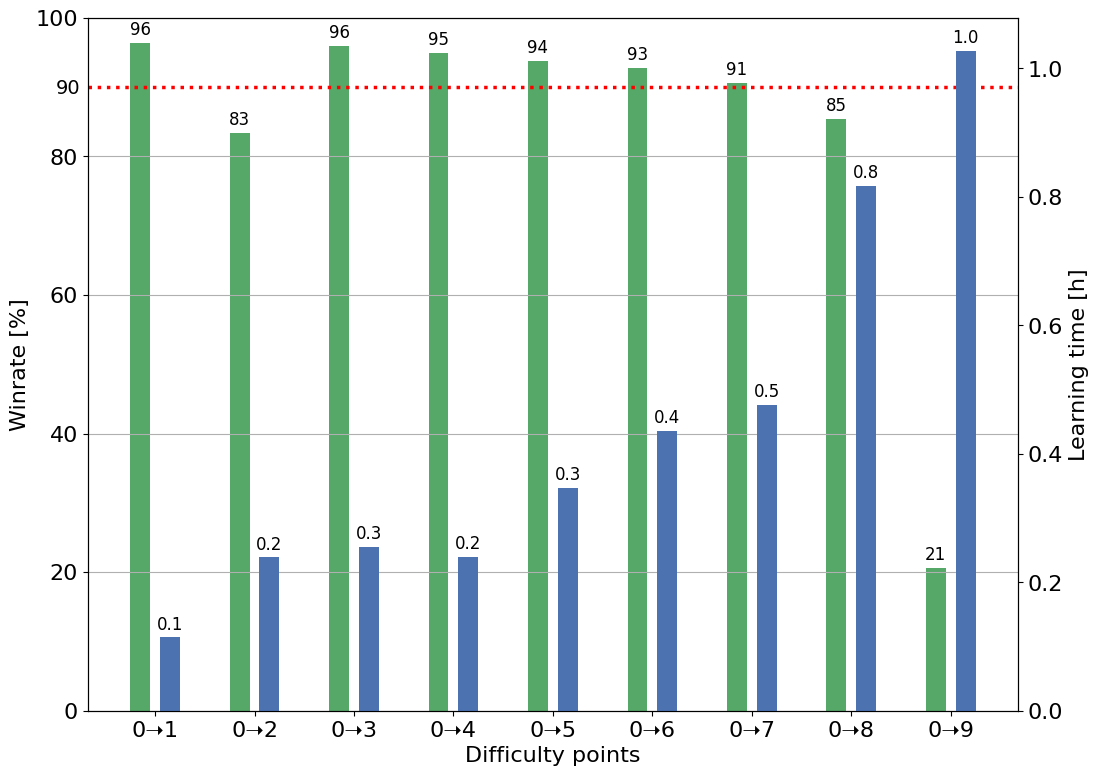}
b)\includegraphics[width=0.8\linewidth]{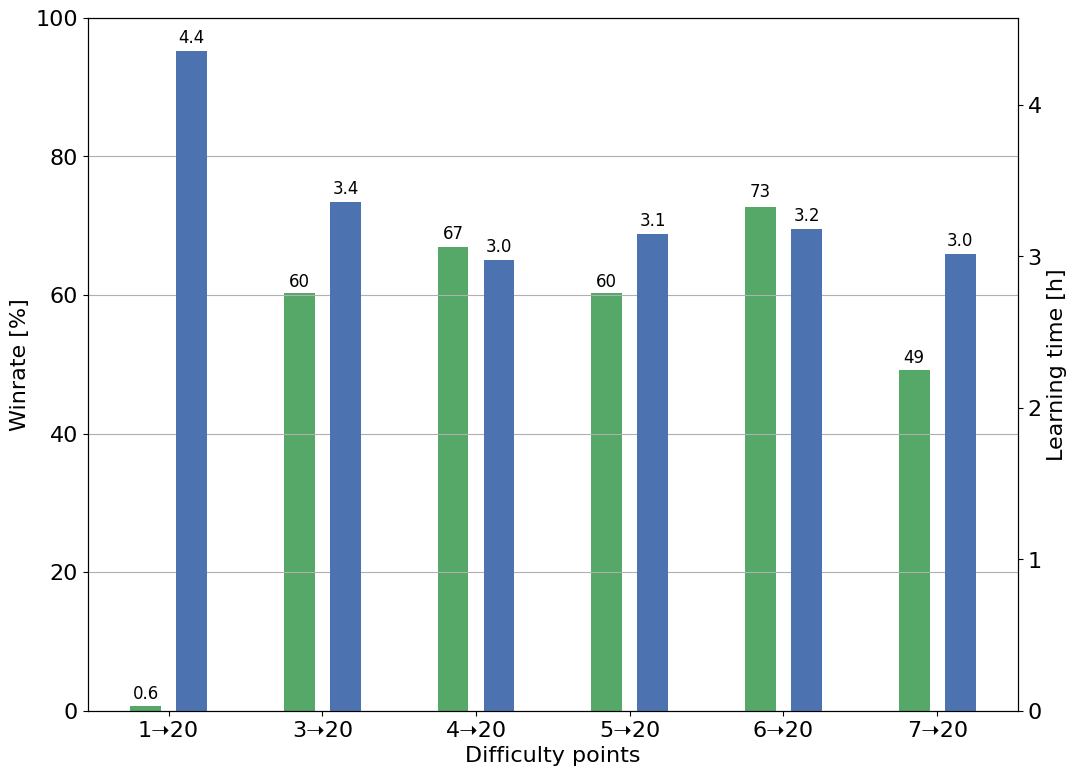}
\caption{Distribution of winrate for a) first step of learning (difficulty from 1 to 9), b) second step of learning at full difficulty (20 points).}
\label{fig:select_by_winrate}
\end{figure}

The first solution analysed was to select only a few networks for learning on the game with simplified difficulty. Figure~\ref{fig:select_by_winrate} shows the distribution of the win rate after testing the learning networks in two steps. Chart~(a) is for the first learning step on games of reduced difficulty (from 1 to 9 points), while the second (b) is the results after learning the selected networks on a game of difficulty 20. 

Only networks whose winrate was above 90\% were admitted to the second step. Therefore, only six cases are analysed in the second stage. The best final result was achieved by a network that learned the $0\rightarrow6\rightarrow20$ scheme, with a probability of victory of 73\%.

The right axis in Figure~\ref{fig:select_by_winrate} shows the time of the learning process. For the first graph (a), it increases with the difficulty of the game. This effect is related to the increasing number of rounds required to win the game at a more difficult level. Learning in the second step (b) does not show this trend, but the increased computational budget (20x2500 episodes) increased one learning process to more than 3 hours.

%%%%%%%%%%%%%%%%%%%%%%%%%%%%%%%%%%%%%%%%%%
\subsection{Two-step interrupted learning}

With the assumed calculation budgets, a full calculation in a two-step scheme takes about 24 hours . The question therefore arose as to whether it would be possible to reduce this time and what impact this would have on the quality of the solution. 

A simple reduction in the number of episodes and iterations of the learning process immediately resulted in a lower win rate. A scheme was therefore proposed in which network testing was abandoned after the first step and selection was based on exceeding the averaged reward during learning. Final testing on the full game difficulty was only conducted at the end. 

The algorithm allows a smooth setting of the reward threshold, after which learning will be interrupted. This way, the number of instances selected for the second learning step can be controlled.

\begin{figure}[htbp]
\centering
a)\includegraphics[width=0.8\linewidth]{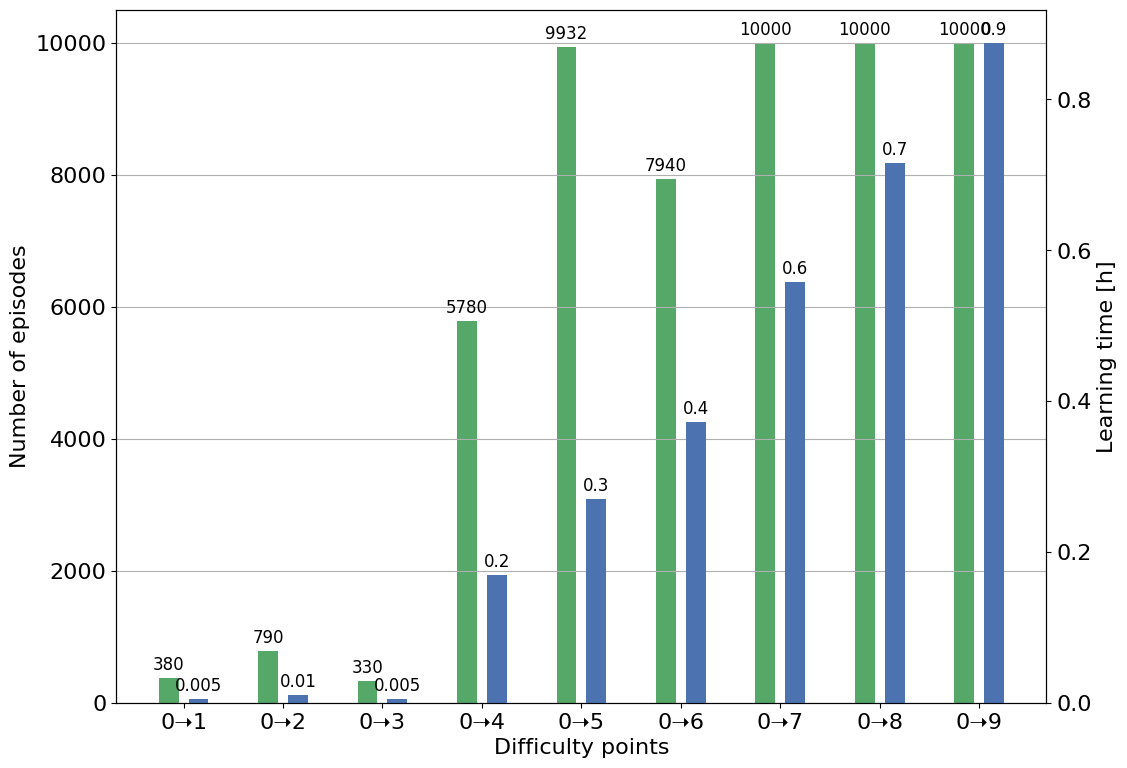}
b)\includegraphics[width=0.8\linewidth]{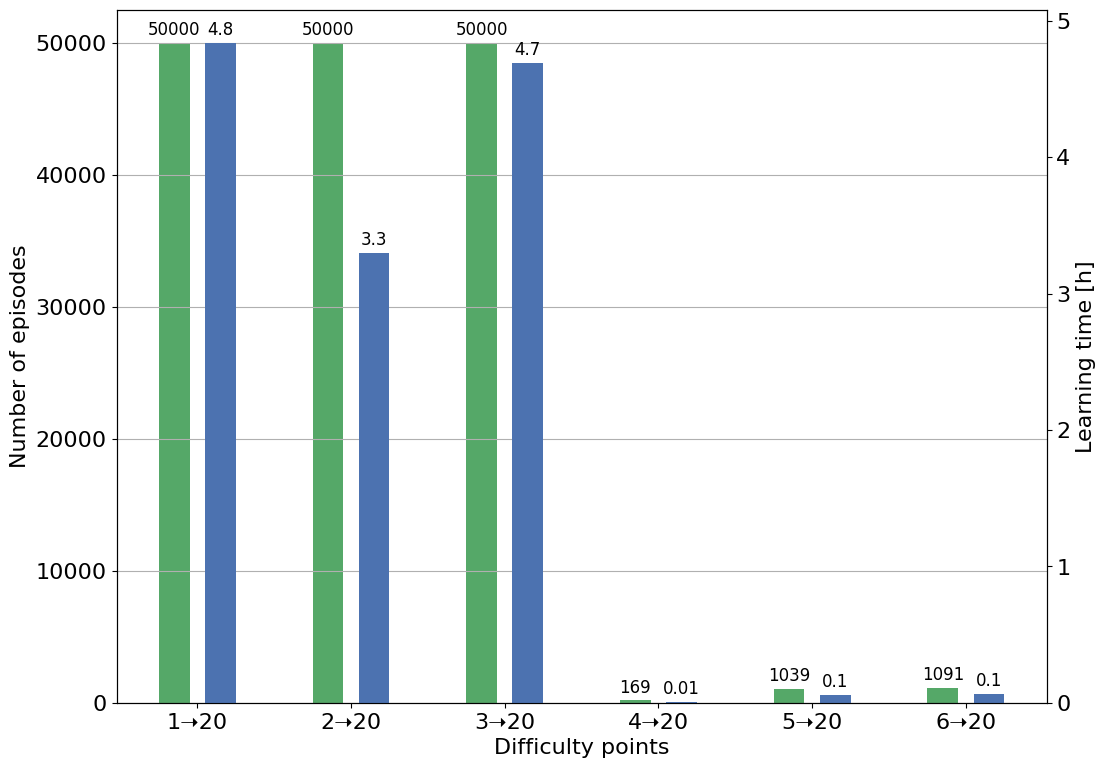}
\caption{Number of episodes and learning time for first(a) and second b) step in interrupted learning scheme. The average reward threshold is: a) greater than 0.5, b) greater than -0.1.}
\label{fig:select_by_threshold}
\end{figure}

Figure~\ref{fig:select_by_threshold}a shows that for a game with low difficulty levels (1-3), the model needs to play less than 1000 games to reach the expected threshold of average reward. For higher difficulties, this number increases until the learning process breaks the budget limit of 10,000 episodes. As in the previous experiment, a strong advantage of low computation time for small values of difficulty points can be seen.

On this basis, six networks were selected for the second step out of the nine tested. The second learning step was carried out in a similar way as before. To improve the quality of the solution, the budget was increased to 20 iterations of the learning process, each consisting of 2,500 episodes. Learning was carried out on a game with a target difficulty of 20 points. As seen in Fig.~\ref{fig:select_by_threshold}b the networks learned on difficulties 1-3 could not fully cope with the new task, and the learning process proceeded until the 50,00 episode limit. In contrast, the networks learned in the first step on difficulties 3-6 also learned relatively quickly to win on the full difficulty. Final testing confirmed that the best result was achieved by the $0\rightarrow6\rightarrow20$ scheme, giving a winning factor of 64\% after 16 hours of computation.

\subsection{Strategies comparison}

\begin{figure*}
\centering
\includegraphics[width=0.8\textwidth]{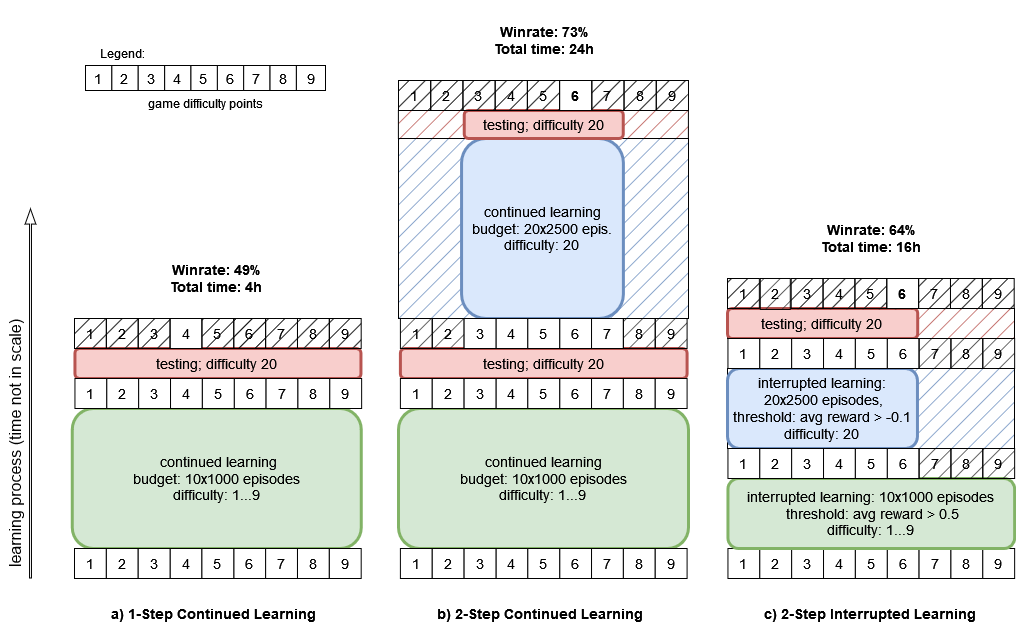}
\caption{Illustration of three examined learning strategies. The simplest is a) one-step continued process, where the agent is learning 9 times for different difficulties. The second scheme b) involves two step learning without interruption. The third is strategy c) where learning is interrupted when the average reward reaches the set threshold. }
\label{fig:transfer_learning}
\end{figure*}

Graphical comparison between the three described strategies of learning is presented in Fig.~\ref{fig:transfer_learning}.

The research aimed to determine the ideal difficulty level for the learning process in the game.  Analyses were conducted for each strategy from difficulties 1 to 9, concluding with testing at level 20. 

The research aimed to determine the ideal difficulty level for the learning process in the game. The illustration highlights a unique internal learning architecture that utilises successive modules, represented by coloured rectangles, operating on results from the previous module. The computation time elapsed is shown on the vertical axis. To enhance readability, the graph does not preserve proportions, so the total time is indicated on the top of each strategy. The most significant computational effort takes place during the second learning step, in which up to 50,000 games were played at the highest difficulty level 20. 

The eliminated cases are represented by the crossed-out elements on the graph. By selection, we cut down the number of analyzed alternatives, accelerating the learning process. Eliminating analyses in the second stage of learning is especially crucial due to their high computational weight.

Fig. ~\ref{fig:transfer_learning} clearly shows that difficulty 6 is the optimal midpoint when learning in two steps. Interrupted learning allowed to find the same solution, in less time. It would therefore be possible to use strategy (c) to find the optimal midpoint, and then strategy (b) already for the chosen optimal difficulty 6 only.

Experiments were also carried out with dividing the learning into more steps, but these did not give a better quality solution, and resulted in longer computation times. We conclude that in the LOTRCG game two learning steps are the optimal solution.

\section{Multi-agent setup}

The analyses presented so far have dealt with the use of a single RL agent to make decisions in the questing phase. This was an initial simplification, but it should be remembered that each round of the game contains five decision moments. In previous research \cite{PhDKonrad} we identified three of them (planning, questing and defence phase in Fig.~\ref{fig:phase_seq}) as the most important.

\begin{table}
\centering
\caption{Comparison of winrates for different setups of agents.  }
\begin{tabular}{ccc|c}
\toprule
\multicolumn{3}{c|}{AI setup} & winrate \\
\hline
planning & questing & defense &  \\
\midrule
RL & random & random & $11.2 \pm 0.6$ \\
random & RL & random & $28.3 \pm 0.9$ \\
random & random & RL & $7.0 \pm 0.5$ \\
\hline
RL & RL & random &  $66.2 \pm 0.9$ \\
RL & random & RL &  $16.3 \pm 0.7$ \\
random & RL & RL &  $33.6 \pm 0.9$ \\
\hline
RL & RL & RL &  $64.4 \pm 0.9$ \\
\bottomrule
\end{tabular}
\label{tab:macroAC}
\end{table}

Every of those three decisions could be managed by RL agents. At the same time, it must be remembered that any increased number of RL agents increases the computation time. It is also possible to combine RL and random agents, each specialising in a different decision. Table~\ref{tab:macroAC} presents comparison for different number of RL agents used. With only one RL agent, the best solutions can be seen when using it at the questing phase (28\%). Two agents on the planning and questing phases yields a result (66\%) that is significantly better than for only one agent. The simultaneous combination of three RL agents (winrate 64\%) does not significantly change the quality of the solution.  

\begin{table}
    \centering
    \caption{Comparison between one RL agent setup vs. two RL agents - final winrates in testing and learning time. }
    % \resizebox{\columnwidth}{!}{
    \begin{tabular}{c|cc|}
         & \multicolumn{2}{c}{random-RL-random}  \\
         & winrate [\%] & learning time [h]  \\
         \hline\hline
         1-step cont. learning & 48.8 & 4.0 \\
         \hline
         2-step cont. learning & 72.7 & 24.0 \\
         \hline
         2-step inter. learning & 64.2 & 15.9 \\
    \end{tabular}
    
\vspace{1em}
    \begin{tabular}{c|cc|}   
             & \multicolumn{2}{c}{RL-RL-random}  \\
         & winrate [\%] & learning time [h]  \\
         \hline\hline
         1-step cont. learning & 71.4 & 9.1 \\
         \hline
         2-step cont. learning & 78.5 & 39.4 \\
         \hline
         2-step inter. learning & 72.2 & 19.8 \\
    \end{tabular}
    % }
    \label{tab:winrates_final}
\end{table}

The two-agent configuration appears to be the most economic solution, so further research has been devoted to it. Consequently, all the tests described in Section 3 were repeated. This time with two RL agents, on the planning and questing phases.

Table~\ref{tab:winrates_final} contains final comparison of results obtained by best one RL setup (random-RL-random), and the best two RL combination (RL-RL-random). Results for one RL agent have been already presented on Fig.~\ref{fig:transfer_learning}, where the winner is two-step learning with full budget.  For two RL agents setup all learning strategies provides winrate above 70\%. However, the learning time is clearly higher. In both uninterrupted cases approximately twice as long. 

It is interesting to compare the results of two agents learning in a one-step strategy with an intermittent two-step learning strategy. A similar win rate (approximately 72\%) was achieved in more than twice the time. Thus, the one-step learning scheme should not be abandoned only on simple problems and testing on the most difficult ones. 

The highest score (78.5\%) was achieved, however, after a long 39-hour uninterrupted, dual-agent RL learning process, which was broken down into two steps. 
In the first step, a neural network with random weights was learning to play at difficulty 6, and then the network was subjected to a learning process at maximum difficulty 20 points.

\section{Conclusions}
It has been demonstrated that Reinforcement Learning techniques can be utilised to construct an agent that dominates in the sophisticated and strategic card game Lord of The Rings. The game is characterised by multiple stages featuring five decision-making phases alongside random events and rule-based actions.

The study has indicated that the quality of the result is heavily influenced by the patterns used in game state coding and action decoding. Furthermore, as anticipated, tuning the hyperparameters of the artificial neural network resulted in a noticeable increase in the average percentage of the winrate.

Much of the research was dedicated to discovering techniques for model learning. This was necessary as direct learning the game at its highest difficulty level was unfeasible. Three approaches were trialled, employing regulated alterations in gameplay complexity and interrupting the learning process. Results indicated that the most effective method was two-step learning without interruptions. However, this was found to require a great deal of computational power. Interruption schemes based on an estimated average reward threshold provide a more cost-effective solution.  

The analysed game features decisions of varying nature, prompting the need to merge agents specialised in different phases. The research identified planning and questing as the crucial phases. Utilising a two-step learning strategy, a model was developed that attained a 78.5\% winrate when tested on 10,000 random games at the highest difficulty level. This is significantly better than the previous studies on the use of MCTS methods\cite{godlewski2021optimisation}, which achieved an average winrate of 40\%, as well as 60\% reported for a similar collectible card game~\cite{miernik2021evolving}.

It is possible to enhance the learning strategy of a team of collaborating agents. Learning each agent separately and subsequently training them in collaboration would be beneficial to achieve this computationally. Moreover, independent agents should be able to communicate with each other through the development of further encodings. This structure could be hierarchical or employ a combination of various AI algorithms (e.g. Multilayer Perceptron, Long-Short Term Memory, Transformer), which showed to be impressively efficient for Starcraft II game~\cite{vinyals2019grandmaster}. 

\bibliographystyle{plain} % We choose the "plain" reference style
\bibliography{refs}

\end{document}